# Sensitivity Analysis on Transferred Neural Architectures of BERT and GPT-2 for Financial Sentiment Analysis

Andy Xie[1]
ruoc.xie@mail.utoronto.ca

Camille Bruckmann[1]
camille.bruckmann@mail.utoronto.ca

Tracy Qian[1]
tracy.qian@mail.utoronto.ca

## Abstract

The explosion in novel NLP word embedding and deep learning techniques has induced significant endeavors into potential applications. One of these directions is in the financial sector. Although there is a lot of work done in state-of-the-art models like GPT and BERT, there are relatively few works on how well these methods perform through fine-tuning after being pre-trained, as well as info on how sensitive their parameters are. We investigate the performance and sensitivity of transferred neural architectures from pre-trained GPT-2 and BERT models. We test the fine-tuning performance based on freezing transformer layers, batch size, and learning rate. We find that BERT's parameters are hypersensitive to stochasticity in fine-tuning and that GPT-2 is more stable in such practice. It is also clear that the earlier layers of GPT-2 and BERT contain essential word pattern information that should be maintained.

## 1 Background

### 1.1 Introduction

The introduction of transformer-based architectures and attention heads in NLP has led to many breakthroughs in a range of tasks, the most prominent being GPT and BERT [1, 2]. One popular task is sentiment analysis [3]. Depending on whether new information received is good or bad, analysts can make better predictions on financial sentiment based on sentiment alone. However, there is little work on how well the dominant architectures will perform given different setup and an imbalanced dataset, which is common in industry. This paper explores the performance of transferred pre-trained GPT-2 and BERT architectures in sentiment classification across 2 benchmarks. Code can be found at https://github.com/axie123/gpt_bert_transfer_arch.

### 1.2 BERT

BERT or Bidirectional Encoder from Transformers (BERT), consists of stacking encoders to form 12 transformer layers. BERT is able to solve problems such as neural machine translation, question answering, and sentiment analysis [1]. In order to solve these problems, BERT must be trained in two stages: pre-training and fine-tuning [1]. During pre-training, the goal is for BERT to understand language and context. It achieves this by training on two unsupervised tasks simultaneously: Masked Language Modeling (MLM) and Next Sentence Prediction (NSP) [1]. MLM helps BERT to understand bi-directional context within a sentence [1]. The goal of NSP is to classify whether one sentence follows another, allowing BERT to build an understanding of long-term dependencies across sentences.

---
[1] University of Toronto

## 1.3 GPT-2

GPT2 is an enhanced version of the Generative Pre-trained Model (GPT) created by OpenAI [2]. Classified as artificial general intelligence (AGI), GPT2 has the ability to perform a variety of different tasks, ranging from text summarization to question answering, without being specifically trained to do so [2]. GPT2's architecture is a 12-layer, decoder-only transformer using twelve masked causal self-attention heads [2]. Similar to BERT, training GPT2 requires a pre-training and a fine-tuning phase.

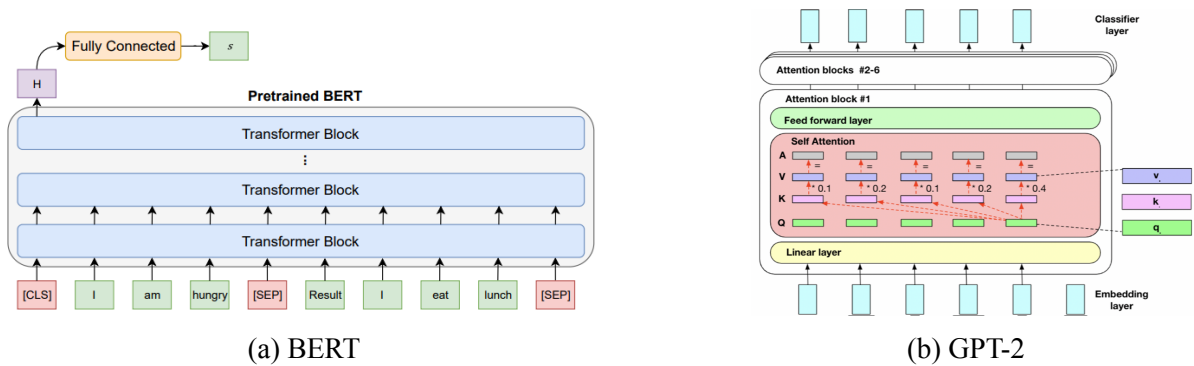

(a) BERT  (b) GPT-2

Figure 1: (a) BERT Sequence Classifier, (b) GPT-2 Self-Attention Classifier [4, 5]

# 2 Benchmarks

We will be using the FiQA and Financial PhraseBank datasets for this investigation [6, 7]. The combined benchmarks contains 5842 parsed financial news sentences with three classes: positive, neutral and negative. These labels were tokenized into indices for classification. We were able to find a pre-labeled CSV file of the desired benchmarks combined from Kaggle [8]. To tokenize our training and testing sentences, we used BERT and GPT2 fast tokenizer from Hugging Face [9, 10]. The training-validation-test split is 70%-15%-15%.

# 3 Model Configurations

The specific pretrained models used for the experiment were the GPT-2 and BERT variations for sequence classification [9, 10]. These two models are the classification variations of the base transformer models with a linear classification head as the output layer, whose size is equivalent to the number of unique labels in the dataset. A weighted cross entropy loss is used to make up for the imbalanced dataset.

The baseline configurations are listed in the table below. All models use a weight decay rate of 0.01. The activation function for the attention heads in each layer is SELU. Both models were trained for 40 epochs.

| Baseline Model | Active Layers | Batch Size | Learning Rate | Test Accuracy | F1 Scores |
| GPT-2 | 12 | 64 | 0.001 | 70.53 | 0.75, 0.79, 0.36 |
| BERT | 12 | 32 | 0.0005 | 53.59 | 0.70, 0.00, 0.00 |

Table 1: Baseline Model Performance of GPT-2 and BERT.

## 3.1 Effect of Transformer Layer Deactivation

For each transformer model, we decided to freeze certain transformer layers to deactivate them before fine-tuning. The same hyperparameter values from their respective baselines are used for this section. All model variations with frozen layers are trained for 40 epochs. The validation accuracies for GPT-2 and BERT models with frozen layers are displayed in Appendix 1 and 2.

| Model | Frozen Layers 1-2 | Frozen Layers 1-4 | Frozen Layers 1-6 | Frozen Layers 1-8 | Frozen Layers 1-10 |
|---|---|---|---|---|---|
| GPT-2 | 0.75, 0.82, 0.37 | 0.76, 0.84, 0.41 | 0.77, 0.86, 0.42 | 0.79, 0.87, 0.45 | 0.78, 0.86, 0.44 |
| BERT | 0.70, 0.00, 0.00 | 0.70, 0.01, 0.00 | 0.70, 0.00, 0.00 | 0.79, 0.87, 0.49 | 0.79, 0.87, 0.49 |

Table 2: F1-scores of test set for layer freezing, each cell reports neutral, positive, and negative labels.

We observe that the performance of GPT-2 improves gradually when more layers are frozen, and peaks off with 1-6 and 1-8 layers frozen. However, the improvements are insignificant. In contrast, BERT experiences degeneracy when transformer layers 1-6 are active. It can be said that the early self-attention layers in GPT-2 are able to adapt much better to different classification tasks during fine-tuning than those of BERT.

## 3.2 Effect of Different Batch Sizes

GPT-2 with frozen layers 1-6 and BERT with frozen layers 1-8 are used as baselines configurations to investigate the effect of different batch sizes on both language models. Both batch size and weight decay are maintained from the baselines. The models are trained for 30 epochs. The validation accuracies can be seen in Appendix 3 and 4.

| Model | 8 | 16 | 32 | 64 | 128 |
|---|---|---|---|---|---|
| GPT-2 | 0.77, 0.84, 0.43 | 0.78, 0.83, 0.45 | 0.77, 0.86, 0.42 | 0.77, 0.86, 0.42 | 0.79, 0.87, 0.48 |
| BERT | 0.70, 0.00, 0.00 | 0.70, 0.00, 0.00 | 0.70, 0.00, 0.00 | 0.80, 0.86, 0.53 | 0.79, 0.87, 0.48 |

Table 3: F1-scores of test set for batch sizes, each cell reports for neutral, positive, and negative labels.

The performance of GPT-2 isn't notably affected by the changing of the batch sizes, and the best performing variation of the model (128) only slightly outperforms the others by 0.01. However, BERT shows degeneracy with the batch size being below 64, with only 64 and 128 performing close to GPT-2.

## 3.3 Effect of Learning Rate Changes

This section examines how learning rate affects the fine-tuning efficiency and performance of the model. Seven learning rates are used to fine tune each transformer model. The same numbers of frozen layers are used, with GPT-2 using a batch size of 128 and BERT using a batch size of 64. Each model setup is trained for 30 epochs. The graphs in Appendix 5 and 6 are the validation accuracies.

| Model | 0.01 | 0.005 | 0.001 | 0.0005 | 0.0001 | 0.00005 | 0.00003 |
|---|---|---|---|---|---|---|---|
| GPT-2 | 0.72, 0.67, 0.31 | 0.75, 0.80, 0.43 | 0.80, 0.88, 0.47 | 0.80, 0.87, 0.47 | 0.80, 0.85, 0.50 | 0.80, 0.85, 0.51 | 0.81, 0.85, 0.58 |
| BERT | 0.00, 0.00, 0.26 | 0.70, 0.00, 0.00 | 0.00, 0.48, 0.00 | 0.78, 0.86, 0.52 | 0.79, 0.88, 0.47 | 0.79, 0.87, 0.47 | 0.80, 0.87, 0.49 |

Table 4: F1-scores of test set for learning rates, each cell reports neutral, positive, and negative labels.

GPT-2's performance gradually increases as the learning rate gets smaller, reaching the best performance at the smallest learning rate of 0.00003. BERT only performs well with learning rates smaller or equal to 0.0005. However, we see that both models didn't handle the imbalance of the dataset well across all experiments. With the negative labels being the worst performing despite the weighted loss function.

### 3.4 BERT Extension Fine-Tuning

Unlike GPT-2, we find that BERT's fine-tuning performance is extremely sensitive to batch size and learning rate. So far, we have shown evidence that the knowledge gained by BERT during pre-training can only be adjusted for the classification task by small iterations of GD with large batch sizes. We prove this with the data below. We see only a small batch size and learning rate will lead to similar performance, as a this would likely introduce similar stochasticity in updates. whereas a setup that introduces more stochasticity will lead to degeneracy. We can also see this in Appendix 7.

| Batch Size, Learning Rate | Validation Accuracy, Epoch | Test Accuracy | F1 Score (Neutral) | F1 Score (Positive) | F1 Score (Negative) |
|---|---|---|---|---|---|
| (a) 8, 0.00005 | 79.55, 7 | 76.51 | 0.79 | 0.88 | 0.45 |
| (b) 8, 0.01 | 53.64, 5 | 53.59 | 0.70 | 0.00 | 0.00 |
| (c) 128, 0.01 | 73.33, 1 | 53.59 | 0.70 | 0.00 | 0.00 |

Table 5: Extension experiments for BERT Classifier, (a) Low batch size, low learning rate, (b) Low batch size, high learning rate, (c) High batch size, high learning rate.

## 4 Conclusion

The performance of BERT and GPT-2 can vary depending on the batch size, learning rate, and the number of frozen transformer layers. GPT-2 appears to be more robust to fine-tuning, while BERT's performance diminishes drastically with small changes in the hyperparameters. It was determined that the first 1-6 layers in both models contain key information, vital to the model's success. Future experiments to further explore the sensitivity of GPT-2 and BERT to fine-tuning include using different pre-trained models (such as FinBERT[1]), or measuring the performance on classifying sentences into more than three classes and different types of classes. In conclusion, BERT and GPT-2 are both powerful models for sentiment analysis, however, GPT-2 appears to be less sensitive to fine-tuning, rendering it a favorable model for less experienced developers.

---

[1] https://github.com/ProsusAI/finBERT

# Appendix

## 1	GPT-2 Validation Accuracies by Frozen Layers

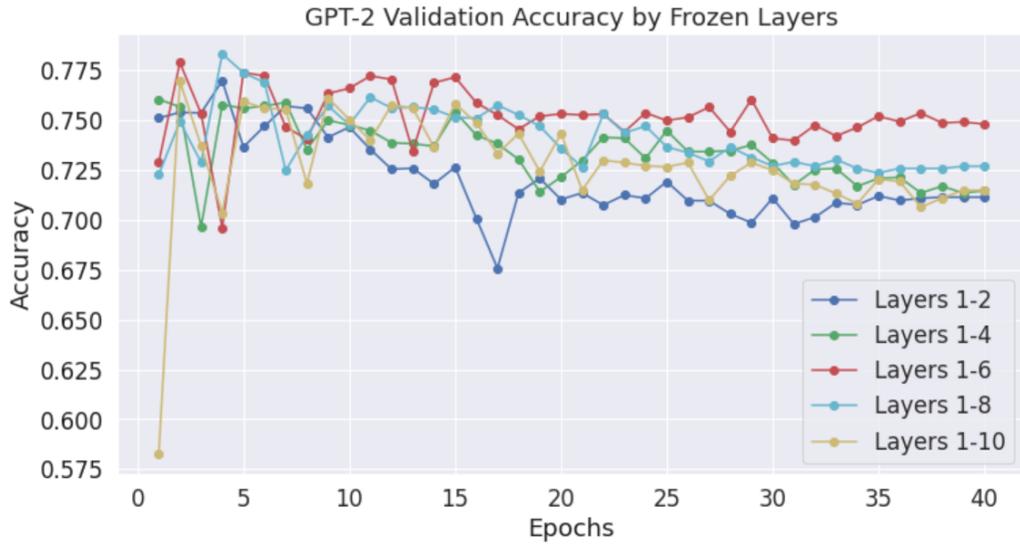

## 2	BERT Validation Accuracies by Frozen Layers

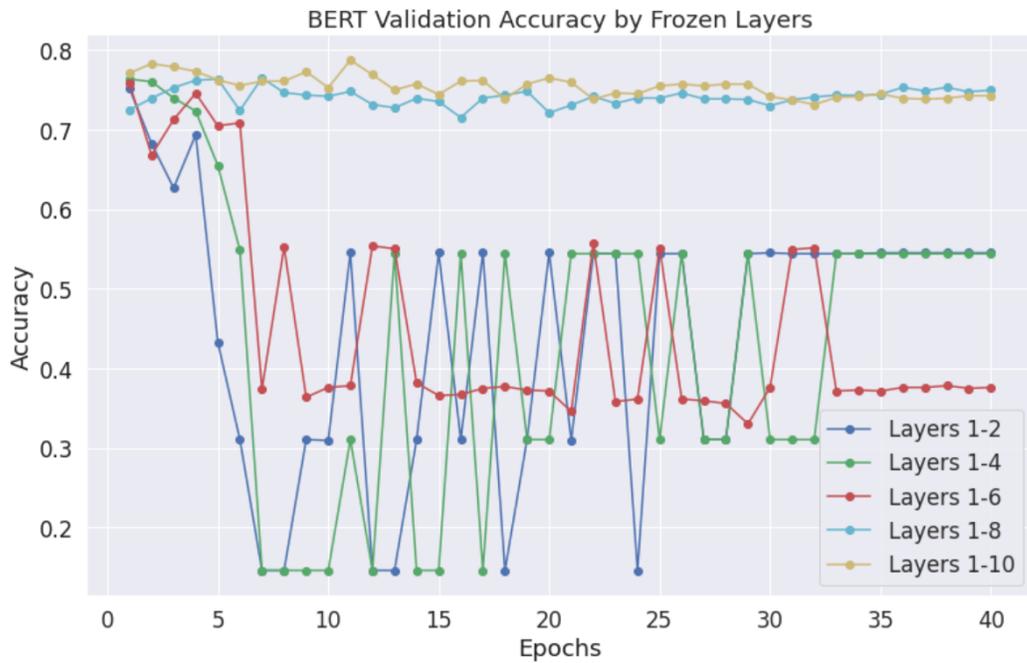

## 3	GPT-2 Validation Accuracies by Different Batch Sizes

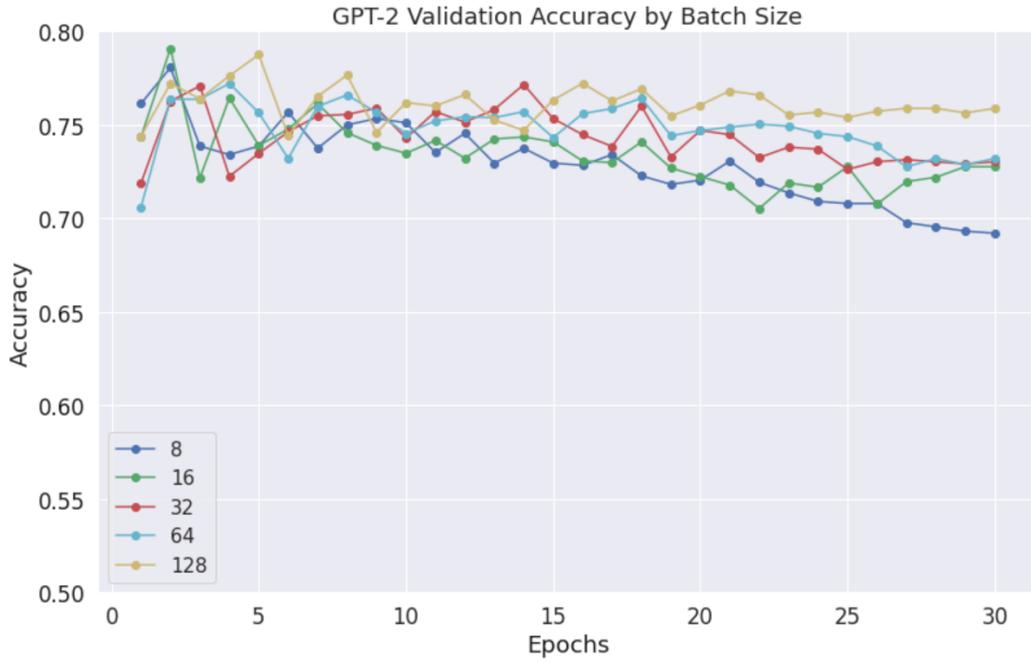

## 4 BERT Validation Accuracies by Different Batch Sizes

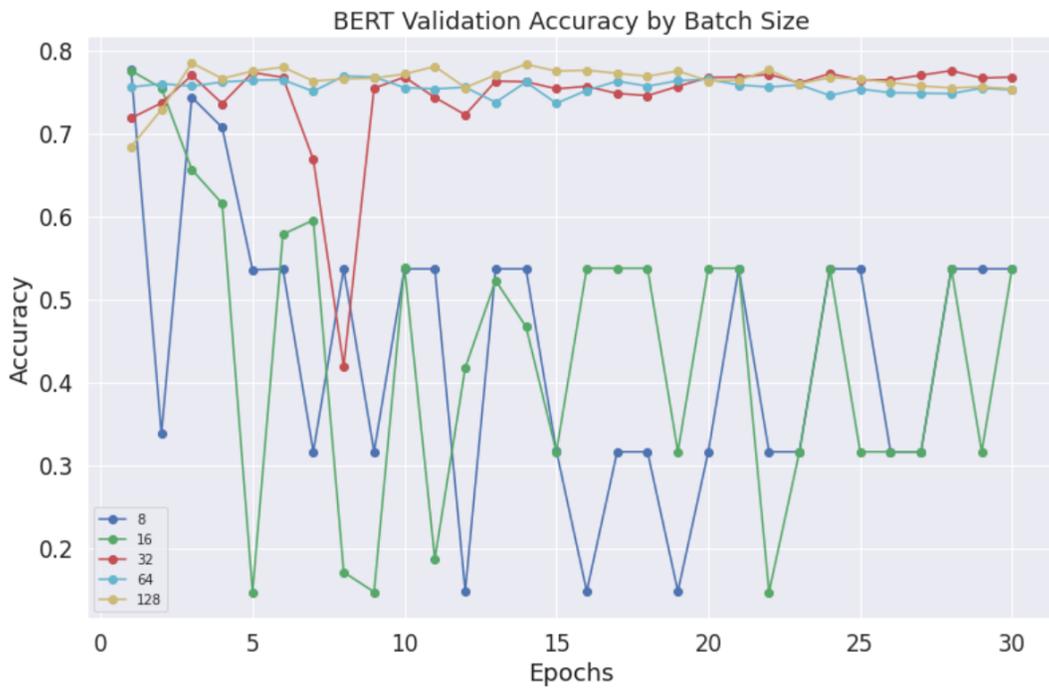

## 5 GPT-2 Validation Accuracies by Different Learning Rates

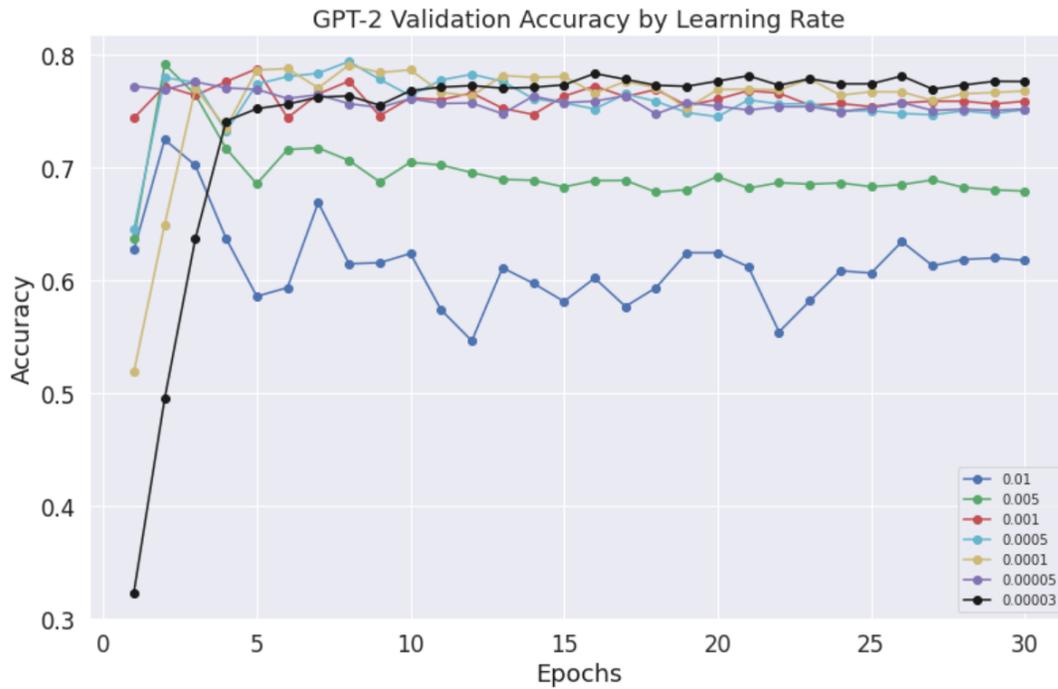

## 6 BERT Validation Accuracies by Different Learning Rates

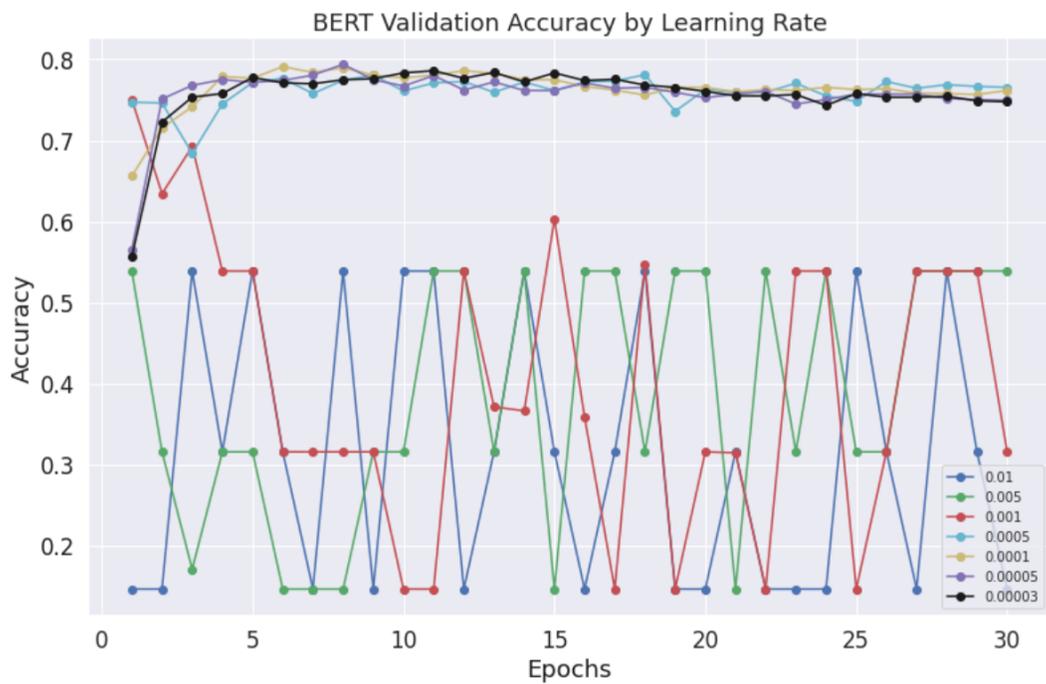

## 7 BERT Extension Experiments

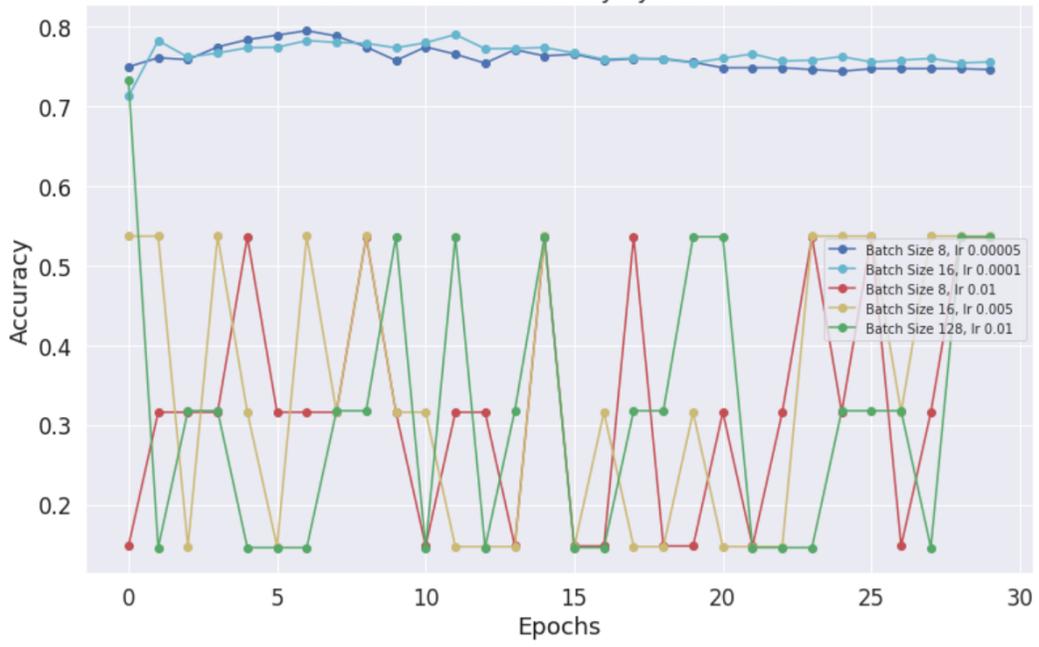